%% file: MWS-SAB-2012.tex
\documentclass{llncs} %
\usepackage{makeidx}  
\usepackage{amsfonts}
\usepackage{graphicx}
\usepackage{algorithmic}
\usepackage{algorithm}

\usepackage{amsmath,amssymb,algorithm,color,graphicx}
\usepackage{wrapfig}
\input{symbols}

\def\trans{^\top\!}
\def\tr{^\top}
\newcommand{\comment}[1]{}
\def\defeq{\buildrel \rm def \over =}
\def\e{{\mathbf e}}
\def\thetaopti{{\theta_*^{i}}}

\begin{document}

\title{Multi-timescale Nexting\\ in a Reinforcement Learning Robot}
\author{Joseph Modayil, Adam White, and Richard S.\ Sutton}
\institute{Reinforcement Learning and Artificial Intelligence Laboratory\\
University of Alberta, Edmonton, Alberta, Canada} 

\maketitle
\begin{abstract}
  The term ``nexting'' has been used by psychologists to refer to the
  propensity of people and many other animals to continually predict
  what will happen next in an immediate, local, and personal sense.
  The ability to ``next'' constitutes a basic kind of awareness and
  knowledge of one's environment.  In this paper we present results
  with a robot that learns to next in real time, predicting thousands of features
  of the world's state, including all sensory inputs, at timescales from 0.1 
  to 8 seconds. This was achieved by treating each state feature as a reward-like target
  and applying temporal-difference methods to learn a corresponding value 
  function with a discount rate corresponding to the timescale.  We show that two thousand predictions, 
  each dependent on six thousand state features, can be learned and 
  updated online at better than 10Hz on a laptop computer, using the
  standard TD($\lambda$) algorithm with linear function approximation.
  We show that this approach is efficient enough to be practical, with
  most of the learning complete within 30 minutes. We also show that a single
  tile-coded feature representation suffices to accurately predict many different signals at 
  a significant range of timescales. Finally, we show that the accuracy of our learned 
  predictions compares favorably with the optimal off-line solution.
\end{abstract}

\section{Multi-timescale Nexting}

Psychologists have noted that people and other animals seem to
continually make large numbers of short-term predictions about their
sensory input (e.g., see Gilbert 2006, Brogden 1939, Pezzulo 2008,
Carlsson et al.\ 2000). When we hear a melody we predict what the next
note will be or when the next downbeat will occur, and are surprised
and interested (or annoyed) when our predictions are disconfirmed
(Huron 2006, Levitin 2006). When we see a bird in
flight, hear our own footsteps, or handle an object, we continually
make and confirm multiple predictions about our sensory input. When we
ride a bike, ski, or rollerblade, we have finely tuned moment-by-moment
predictions of whether we will fall, and of how our trajectory will
change in a turn. 
In all these examples, we continually predict what
will happen to us \textit{next}. Making predictions of this simple,
personal, short-term kind has been called \textit{nexting} (Gilbert,
2006).

Nexting predictions are specific to one individual and to their
personal, immediate sensory signals or state variables. A special name for
these predictions seems appropriate because they are unlike %
predictions of the stock market, of political events, or of
fashion trends. Predictions of such public events seem to involve more
cognition and deliberation, and are fewer in number.
In nexting we envision that one individual may be continually making 
massive numbers of small predictions in parallel. Moreover, nexting predictions
seem to be made simultaneously at multiple time scales. 
When we read, for example, it seems likely that we next at the
letter, word, and sentence levels, each involving substantially
different time scales. 

\comment{
\begin{figure}[tb]
\begin{tabular}{ccc}
\hspace*{-.18in}
\includegraphics[width=.34\columnwidth]{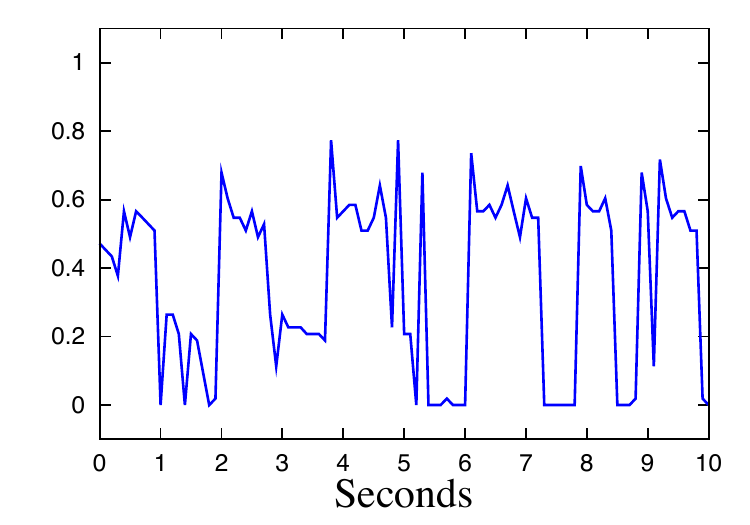} & 
\includegraphics[width=.34\columnwidth]{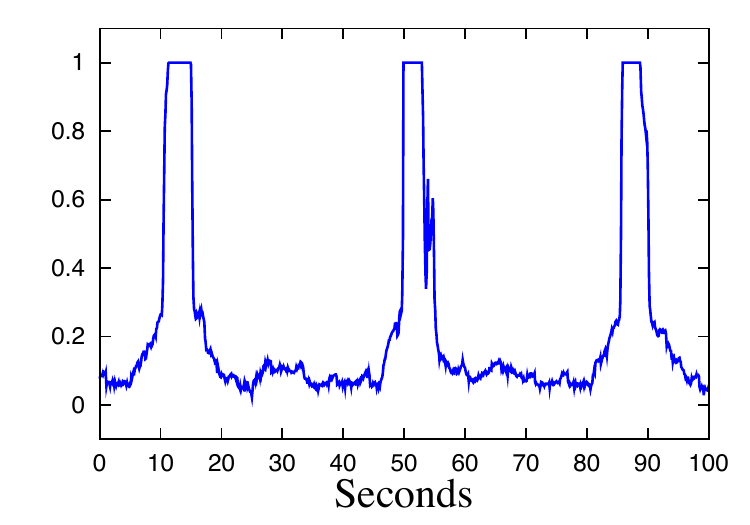} &
\includegraphics[width=.34\columnwidth]{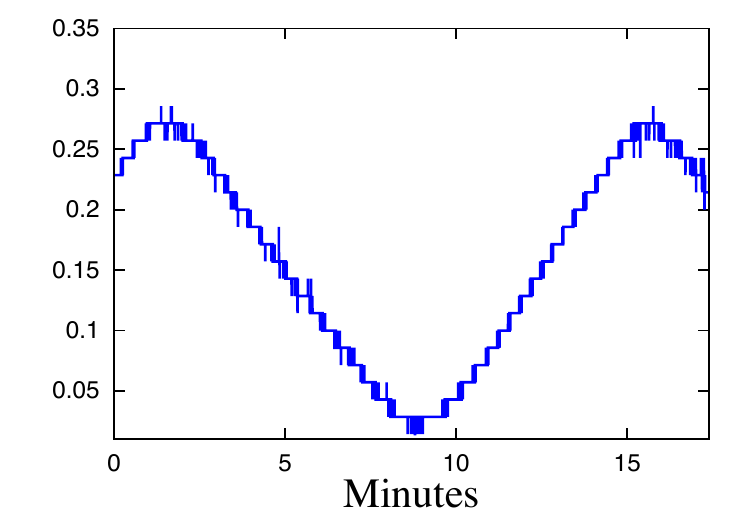} \\
(a) Motor Current & (b) Light & (c) Motor Temperature 
\end{tabular}
\caption{\label{multiscale} Examples of robot sensory signals varying over
different time scales: (a) motor current varying over tenths of a second, 
(b) an ambient light sensor varying over seconds, and (c) a motor temperature sensor
varying over tens of minutes. }
\end{figure}
}

The ability to predict and anticipate has often been proposed as a key
part of intelligence (e.g., see Tolman 1951, Hawkins \& Blakeslee 2004,
Butz et al.\ 2003, Wolpert et al.\ 1995, Clark in press). Nexting can be seen as the
most basic kind of prediction, preceding and possibly underlying all
the others. That people and a wide variety of animals learn and make
simple predictions at a range of short time scales in conditioning
experiments was established so long ago that it is known
as \textit{classical conditioning} (Pavlov 1927). 
Predictions of upcoming shock to a
paw may reveal themselves in limb-retraction attempts a fraction of a
second before the shock, and as increases in heart rate 30 seconds
prior.  In other experiments, for example those known as
\textit{sensory preconditioning} (Brogden 1939, Rescorla 1980), it
has been clearly shown that animals learn predictive relationships
between stimuli even when none of them are inherently good or bad
(like food and shock) or connected to an innate response. In this case
the predictions are made, but not expressed in behaviour until some
later experimental manipulation connects them to a response.  Animals
seem to just be wired to learn the many predictive relationships in their
world.

To be able to next is to have a basic kind of knowledge about how the
world works in interaction with one's body. It is to have a limited form of
forward model of the world's dynamics. To be able to learn to
next---to notice any disconfirmed predictions and continually adjust
your nexting---is to be aware of one's world in a significant way.  Thus, to
build a robot that can do both of these things is a natural
goal for artificial intelligence. Prior attempts to achieve artificial nexting
can be grouped in two approaches. 

The
first approach is to build a \textit{myopic} forward model of the
world's dynamics, either in terms of differential equations or
state-transition probabilities (e.g., Wolpert et al.\ 1995, Grush
2004, Sutton 1990). In this
approach a small number of carefully chosen predictions are made of
selected state variables with a public meaning. The model is myopic in
that the predictions are only short term, either infinitesimally short
in the case of differential equations, or maximally short in the case
of the one-step predictions of Markov models. In these ways, this
approach has ended up in practice being very different from nexting.

The second approach, which we follow here, is to use
temporal-difference (TD) methods to learn long-term predictions
directly. The prior work pursuing this approach has almost all been in
simulation, and has used table-lookup representations and a small
number of predictions (e.g., Sutton 1995, Kaelbling 1993, Singh 1992,
Sutton, Precup \& Singh 1999, Dayan and Hinton 1993).
Sutton et al.\ (2011) showed real-time learning of TD predictions
on a robot, but did not demonstrate the ability to learn many 
predictions in real time or with a single feature representation.

\section{Nexting as Multiple Value Functions}

We take a reinforcement-learning approach to achieving nexting. In reinforcement learning it is commonplace to learn long-term predictions of reward, called \emph{value functions}, and to learn these using temporal-difference (TD) methods such as TD($\lambda$) (Sutton 1988). However, TD($\lambda$) has also been used as a model of classical conditioning, where the predictions are shorter term and where more than one signal might be viewed as a reward (Sutton \& Barto, 1990). Our approach to nexting can be seen as taking this latter approach to the extreme of predicting massive numbers of target signals of all kinds at multiple time scales.

\begin{sloppypar}
We use a notation for our multiple predictions that mirrors---or rather multiplies---that used for conventional value functions. Time is taken to be discrete, $t=1, 2, 3, \ldots$, with each time step corresponding to approximately 0.1 seconds of real time. Our $i$th prediction at time $t$, denoted $v^i_t$, is meant to anticipate the future values of the $i$th prediction's target signal,  $r^i_t$, over a designated time scale given by the discount-rate parameter $\gamma^i$. In our experiments, the target signal $r^i_t$ was either a raw sensory signal or else a component of a state-feature vector (that we will introduce shortly), and the discount-rate parameter was one of four fixed values. The goal of learning is for each prediction to approximately equal the correspondingly discounted sum of the future values of the corresponding target signal:
\begin{equation}
  v^i_t ~~\approx~~ \sum_{k=0}^\infty (\gamma^i)^k r^i_{t+k+1} ~~\defeq~~G^i_t.
\end{equation}
The random quantity $G^i_t$ is known as the \emph{return}.
\end{sloppypar}

We use linear function approximation to form each prediction. That is, we assume that the state of the world at time $t$ is characterized by the feature vector $\phi_t\in\Re^n$, and that all the predictions $v^i_t$ are formed as inner products of $\phi_t$ with the corresponding weight vectors $\theta^i_t$:
\begin{equation}
  v^i_t ~~=~~ \phi_t\tr\theta^i_t ~~\defeq~~ \sum_j \phi_t(j) \theta^i_t(j), 
\end{equation}
where $\phi_t\tr$ denotes the transpose of $\phi_t$ (all vectors are column vectors unless transposed) and $\phi_t(j)$ denotes its $j$th component. The predictions at each time are thus determined by the weight vectors $\theta^i_t$. One natural algorithm for learning the weight vectors is linear TD($\lambda$):
\begin{equation}
  \theta^i_{t+1} = \theta^i_t + \alpha \left(r^i_{t+1} + \gamma^i \phi_{t+1}\tr\theta^i_t - \phi_t\tr\theta^i_t \right) \e^i_t
\end{equation}
where $\alpha>0$ is a step-size parameter and $\e^i_t\in\Re^n$ is an \emph{eligibility trace} vector, initially set to zero and then updated on each step by
\begin{equation}
  \e^i_t = \gamma^i \lambda \e^i_{t-1} + \phi_t, 
\end{equation}
where $\lambda\in[0,1]$ is a trace-decay parameter.

Under common assumptions and a decreasing step-size parameter, TD($\lambda$) with $\lambda=1$ converges asymptotically to the weight vector that minimizes the mean squared error between the prediction and its return. In practice, smaller values of 
$\lambda\in[0,1)$ are almost always used because they can result in significantly faster learning (e.g., see Sutton \& Barto 1998), but the $\lambda=1$ case still provides an important theoretical touchstone. In this case we can define an optimal weight value $\thetaopti$ that minimizes the squared error from the return over the first $N$ predictions:
\begin{equation}
\label{eq:thetastar}
 \thetaopti = \arg\min_{\theta} \sum_{t=1}^{N} \left( \phi_t\trans \theta - G^i_t \right)^2.
\end{equation} 
This  value  can be computed offline by standard
algorithms for solving large least-squares regression problems, and the performance of this offline-optimal value can be compared with that of the weight vectors found online by TD($\lambda$). The offline algorithm is $O(n^3)$ in computation and $O(n^2)$ in memory, and thus is just barely tractable for the cases we consider here, in which $n=6065$. Nevertheless, $\thetaopti$ provides an important performance standard in that it provides an upper limit on one measure of the quality of the predictions found by learning. This upper limit is determined not by any learning algorithm, but by the feature representation. As we will see, even the predictions due to $\thetaopti$ will have residual error. Thus, this analysis provides a method for
determining when performance can be improved with more experience and
when performance improvements require a better representation.  Note
that this technique is applicable even when experience is gathered
from the physical world, where no formal notion of state is available.

\section{Experimental Setup}
\begin{figure}[tb]
\includegraphics[width=.5\columnwidth]{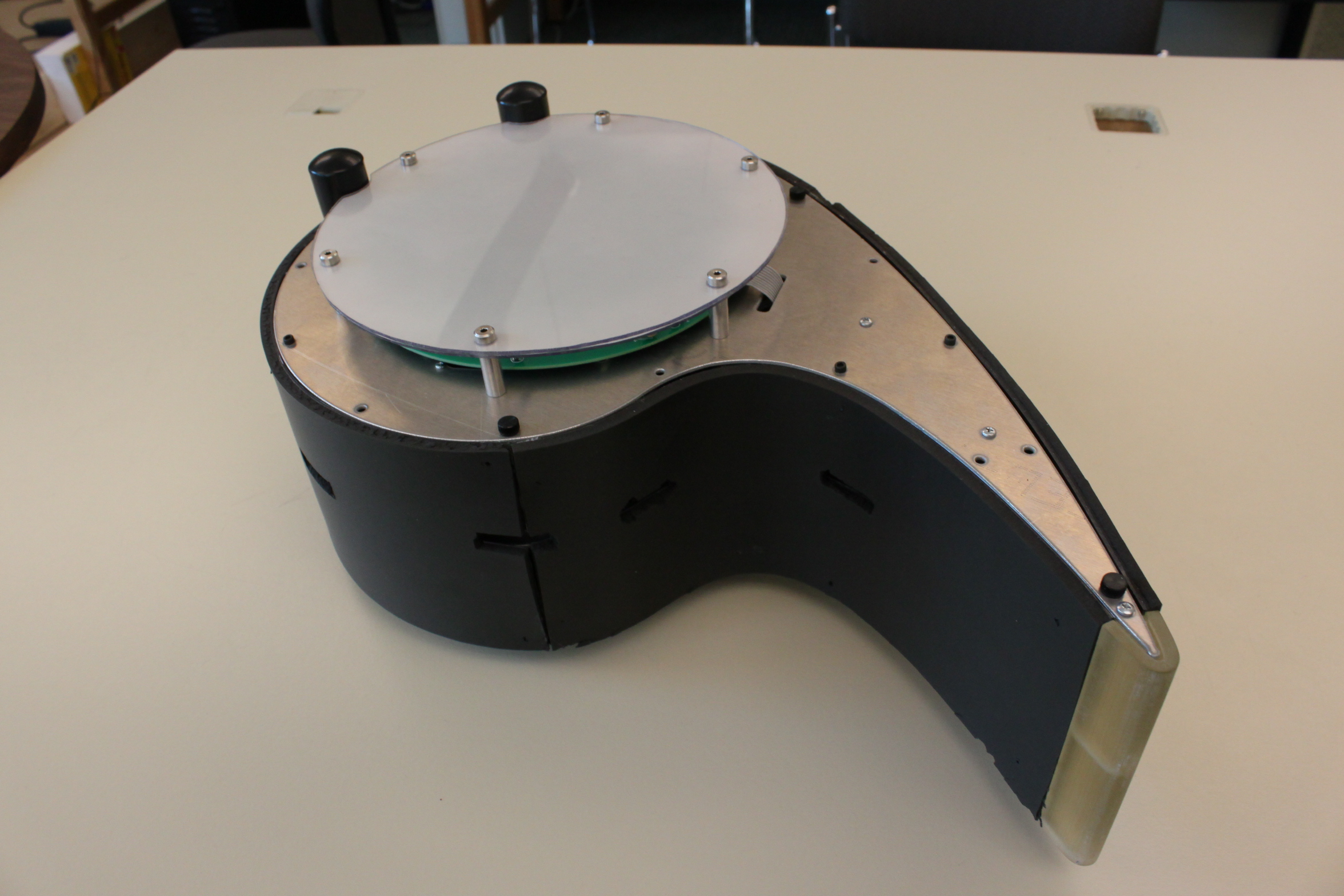}
\includegraphics[width=.45\columnwidth]{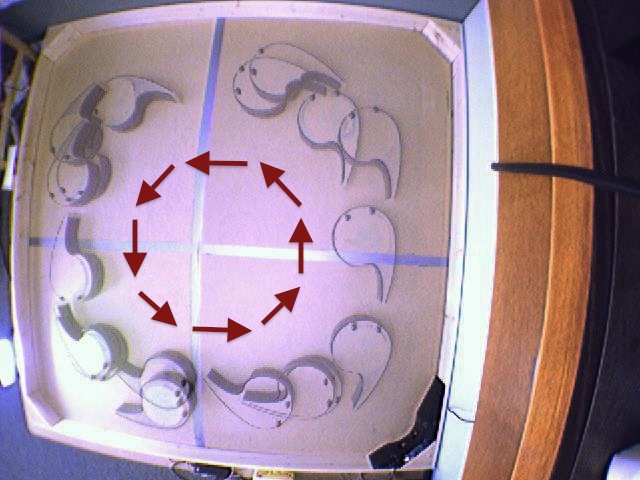}
\caption{\label{hardware} Left: The Critterbot, a
  custom mobile robot with multiple sensors. Right: The Critterbot gathering experience while wall-following in its pen.  This experience contains
  observations of both stochastic events (such as ambient light
  variations from the sun) and regular events (such as
  passing a lamp on the lower-left side of the pen).}
\end{figure}

We investigated the practicality of nexting on the Critterbot, a custom-designed robust and sensor-rich mobile robot platform (Figure~\ref{hardware}, left).  
The
robot has a diverse set of sensors and has holonomic motion provided
by three omni-wheels.  Sensors attached to the motors report the
electrical current, the input motor voltage, motor temperature, wheel
rotational velocities, and an overheating flag, providing substantial
observability of the internal physical state of the robot.  Other
sensors collect information from the external environment.  Passive
sensors detect ambient light in several directions from the top of the
robot in the visible and infrared spectrum. Active sensors emit
infrared light and measure the reflectance, providing information about the distance to nearby obstacles.  Other sensors report acceleration, rotation, and the
magnetic field.  In total, we consider 53 different sensor readings,
all normalized to values between 0 and 1 based on sensor limits.

For our experiments, the agent's state representation was a binary vector, $\phi_t \in \{0,1\}^n$, with a constant number of 1 features, constructed by tile coding  (see Sutton \& Barto 1998).  The features provided no history and performed no averaging of sensor values.  The
sensory signals were partitioned based on sensor modalities. 
Within each sensor modality, each individual sensor (e.g.,
Light0) has multiple overlapping tilings at random offsets (up to 8 tilings), where each tiling splits the sensor range into disjoint intervals of fixed width (up to 8 intervals).
Additionally, pairs of
sensors within a sensor modality 
 were tiled together
using multiple two-dimensional overlapping grids. 
Pairs of sensors were jointly tiled if they were spatially adjacent on the robot (e.g., IRLight$0$ with IRLight$1$) or if there was a single sensor in between them (e.g., IRDistance1 with IRDistance3, IRDistance2 with IRDistance4, etc.). 
All in all, this tiling scheme produced a feature vector with $n=6065$ components, most of which were 0s, but exactly 457 of which were 1s, including one bias feature that was always 1.

\comment{
\begin{table}
\begin{center}
   \small
   \begin{tabular}{ | l | l | l | l |}
   \hline
   Sensor Group & Group  & Tiling Type & (resolution,  \\
& Size & &  \quad tilings)\\ \hline
   IRDistance & 10 & strip & (8,8)  \\ \cline{3-4}
   &  & strip & (2,4)  \\ \cline{3-4}
   & & skip(0) & (4,4) \\ \cline{3-4}
   &  & skip(1) & (4,4)  \\ \hline

   Light & 4 & strip & (4,8)  \\ \cline{3-4}
   &  & skip(0) & (4,1)  \\ \hline

   IRLight & 8 & strip & (8,6)  \\ \cline{3-4}
   &  & strip & (4,1) \\ \cline{3-4}
   &  & skip(0) & (8,1)  \\ \cline{3-4}
   &  & skip(1) & (8,1)  \\ \hline
   Thermal & 8 & strip & (8,4) \\ \hline

   Rotational Velocity & 1 & strip & (8,8)  \\ \hline

   Mag & 3 & strip & (8,8)  \\ \hline

   Accel & 3 & strip & (8,8)  \\ \hline

   MotorSpeed & 3 & strip & (8,4) \\ \cline{3-4}
    &  & skip(0) & (8,8) \\ \hline

   MotorVoltage & 3 & strip & (8,2) \\ \hline
   MotorCurrent & 3 & strip & (8,2)  \\ \hline
   MotorTemperature & 3 & strip & (4,4)  \\ \hline

   OverheatingFlag & 1 & strip & (2,4)  \\ \hline

   LastAction & 3 & strip & (6,4)  \\ \hline

   \end{tabular}
 \end{center}
  \caption{\label{tiles} Summary of the tile-coding strategy for producing the feature vector from the sensory observations. Sensors values in each group were tiled either singly (strip tilings) or jointly pairwise (skip tilings). The last column indicates how many tilings of each type were done for each sensor or sensor group, and how many intervals (resolution) were involved in each dimension of each tiling. See text for explanation.}\end{table}
}

  The robot experiment was conducted in a square wooden pen,
approximately two meters on a side, with a lamp on one edge (see Figure~\ref{hardware}).  The
robot's actions were selected according to a fixed stochastic wall-following policy.
This policy moved forward by default, slid left or
right to keep a side IRDistance sensor within a bounded range
(50-200), and drove backward while turning when the front IRDistance sensor reported a nearby obstacle.  The robot completed a
loop of the pen approximately once every 40 seconds.  Due to
overheating protection, the motors would stop to cool down  at approximately 14 minute intervals.
To increase the diversity of the data, the policy selected an action
at random with a probability  $p=0.05$.  At every time step (approximately 100ms), 
sensory data was gathered and an action performed.  This simple policy
was sufficient for the robot to reliably follow the wall for hours, even
with overheating interruptions.

The wall-following policy, tile-coding, and the TD($\lambda$) learning algorithm were all implemented in Java and run on a laptop connected to the robot by a dedicated wireless link. The laptop used an Intel Core 2 Duo processor with a 2.4GHz clock cycle, 3MB of shared L3 cache, and 4GB DDR3 RAM. The system garbage collector was called on every time step to reduce variability. Four threads were used for the learning code. For offline analysis, data was also logged to disk for 120000 time steps (3 hours and 20 minutes).

\section{Results} 

We applied TD($\lambda$) to learn 2160 predictions in parallel.  The
first 212 predictions had the target signal, $r^i_t$, set to the
sensor reading of one of the 53 sensors and the discount rate,
$\gamma^i$, set to one of four timescales; the remaining 1948
predictions had the target signal set to one of 487 randomly selected
components of the feature vector and the discount rate set to one of
four timescales. The discount rates were one of the four values in
$\{0, 0.8, 0.95, 0.9875\}$, corresponding to time scales of
approximately 0.1, 0.5, 2, and 8 seconds respectively.
The learning parameters were $\lambda = 0.9$ and $\alpha=0.1 / 457 
(=\mbox{\# of active features})$. The initial weight vector was set to zero.

Our initial performance question was scalability, in particular, whether so many predictions could be made and learned in real time. We found that the total computation time for a cycle under our conditions was 55ms, well within the 100ms duty cycle of the robot. The total memory consumption was 400MB. Note that with faster computers the number of predictions or the size of the weight and feature vectors could be increased at least proportionally. This strategy for nexting should be easily scalable to millions of predictions with foreseeable increases in parallel computing power over the next decade.

\comment{
\begin{figure}
\caption{\label{predictedReturn}  The predictions made by the online learner TD($\lambda$) qualitatively matches the empirical return, rising and falling in advance of the observation.  This graph shows some of the potential of nexting, as the robot is able to learn to predict that the signal will rise well in advance of observed changes in the signal.  Also, we see that the prediction made by the online
  learner TD($\lambda$) has comparable performance to the offline-optimal  prediction computed offline.}
\end{figure}
}

\begin{figure}[h]
\includegraphics[width=.51\columnwidth]{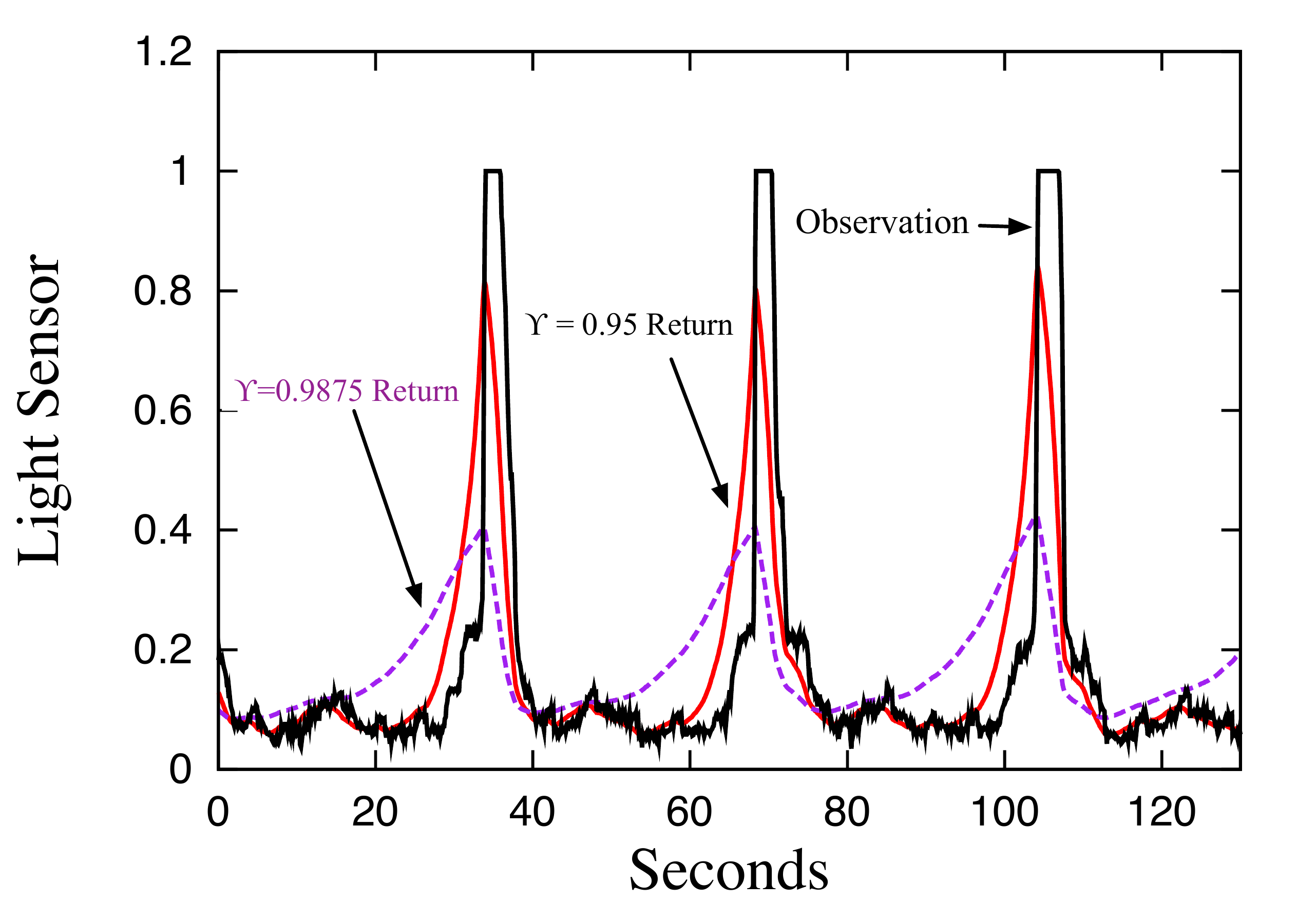}
\includegraphics[width=.51\columnwidth]{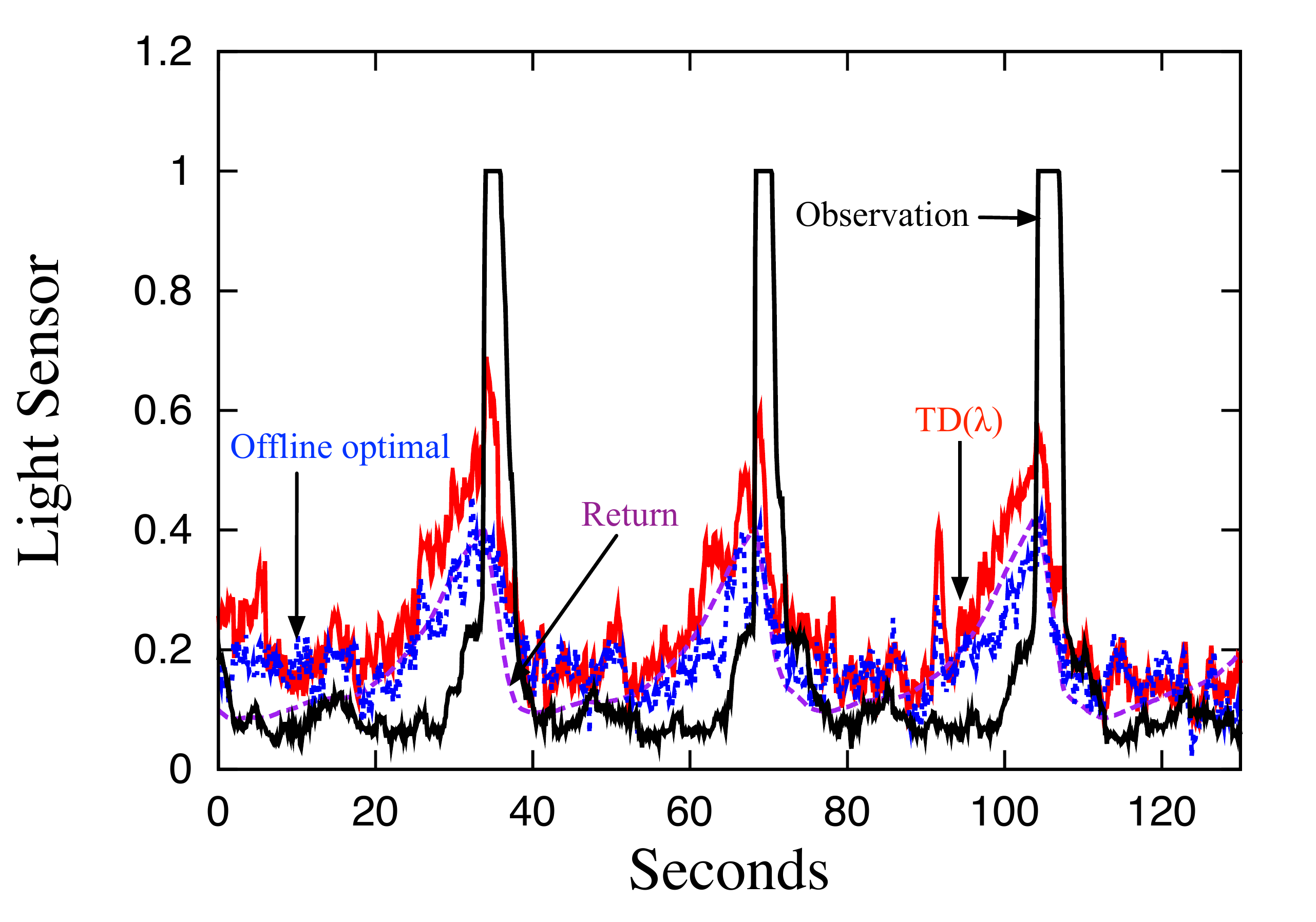}
\caption{\label{empiricalReturn} Nexting is demonstrated in these graphs with predictions that rise and fall prior to the increase and decrease of a sensory signal. Comparison of ideal (left) and learned (right)  predictions of one of the light sensors for three trips around the pen after 2.5 hours of experience. On each trip, the sensor value saturates at 1.0. The returns for the 2 and 8-second predictions, shown on the left, rise in anticipation of the high value, and then fall in anticipation of the low value. The 8-second predictions in the second panel of the offline-optimal weights (dotted blue line) and the TD($\lambda$)-learned weights (solid red line) behave similarly both to each other and to the returns (albeit with more noise). }
\end{figure}

For an initial assesment of accuracy, let us take a close look at one of the predictions, in particular, at the prediction for one of the light sensors. Notice that there is a bright lamp in the lower left corner of the pen in Figure 1 (right). On each trip around the pen, the light sensor increases to its maximal level and then falls back to a low level, as shown by the black line in Figure~\ref{empiricalReturn}. If the state features are sufficiently informative, then the robot may be able to anticipate the rising and falling of this sensor value. The ideal prediction is the return $G^i_t$, shown on the left in the colored lines in Figure 2 for two time scales (two seconds and eight seconds). Of course, to determine these lines, we had to use the future values of the light sensor; the idea here is to approximate these ideal predictions (as in Equation~\ref{eq:thetastar}) using only the sensory information available to the robot in its feature vector.
The second panel of the figure shows the predictions due to the weight vector adapted online by TD($\lambda$) and due to the optimal weight vector, $\thetaopti$, computed offline %
(both for the 8-second time scale). The key result is that the
robot has learned to anticipate both the rise and fall of the light.  Both the
learned prediction and the optimal offline prediction match the
return closely, though with substantial noisy perturbations.

\begin{wrapfigure}{r}{.5\columnwidth}
\vspace*{-20pt}
\includegraphics[width=.53\columnwidth]{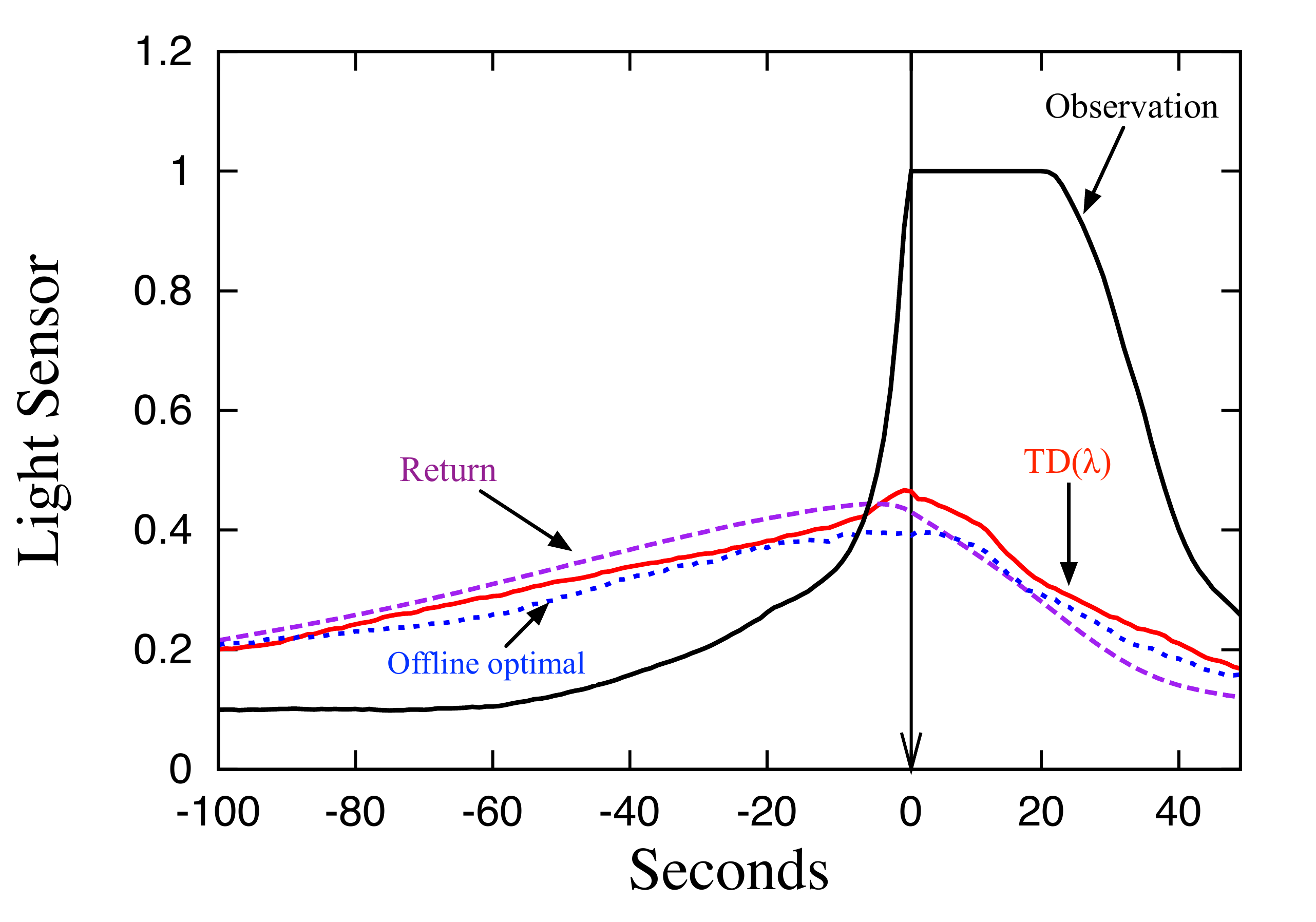}
\vspace*{-20pt}
\caption{\label{alignedReturn} An average of 100 cycles like the three shown in Figure~\ref{empiricalReturn} (right panel), aligned on the onset of sensor saturation.  Error bars are  slightly wider than the lines themselves and overlap substantially, so are omitted for clarity}
\vspace*{-10pt}
\end{wrapfigure}
Figure~\ref{alignedReturn} is a still closer look at this same prediction, obtained by averaging over 100 circuits around the pen, aligning each circuit's data so that the time of initial saturation of the light sensor is the same. We can now see very clearly how the predictions and returns anticipate both the rise and fall of the sensor value, and that both the TD($\lambda$) prediction and the optimal prediction, when averaged, closely match the return.

\begin{figure}[tb]
\includegraphics[width=.5\columnwidth]{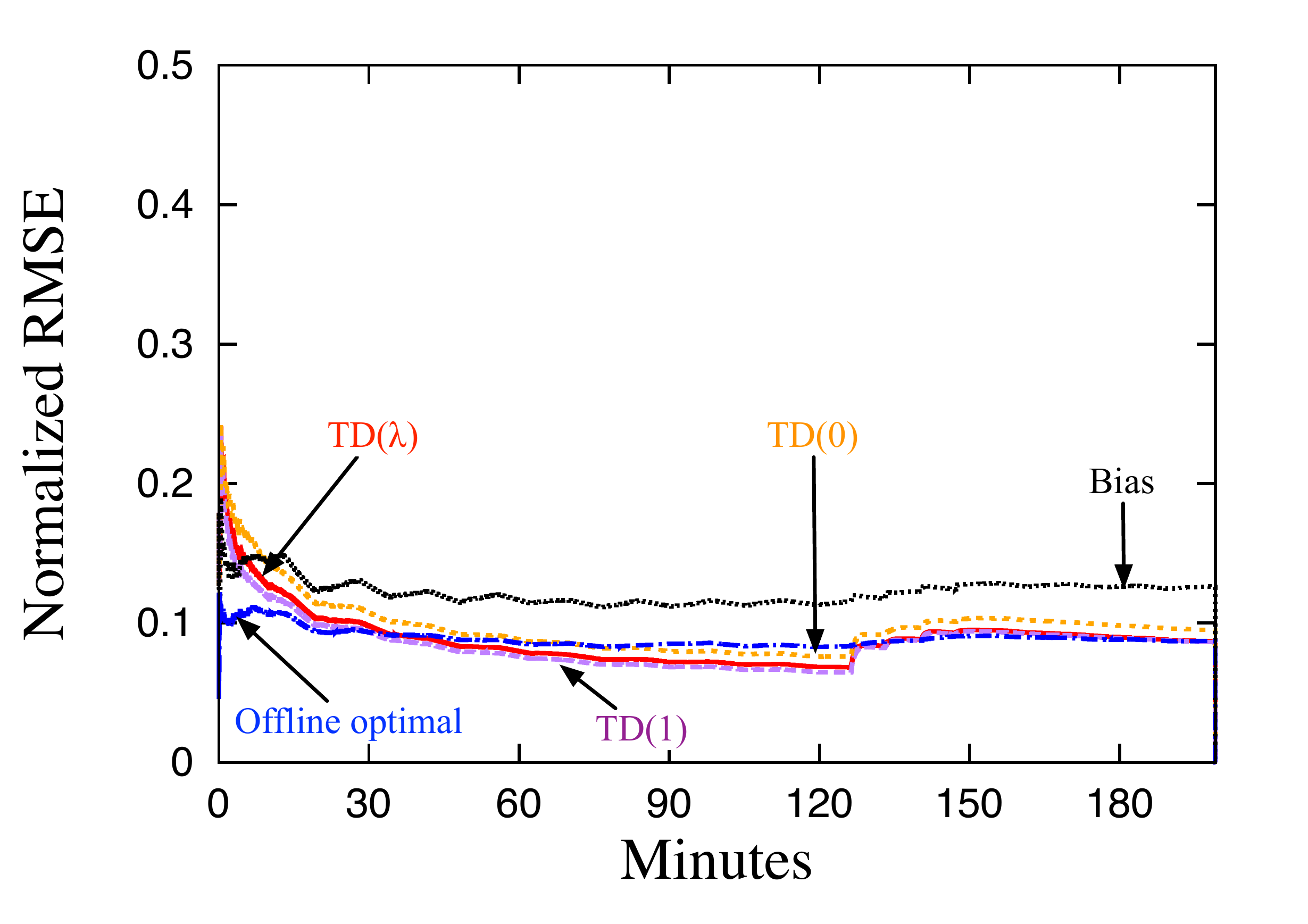}
\includegraphics[width=.5\columnwidth]{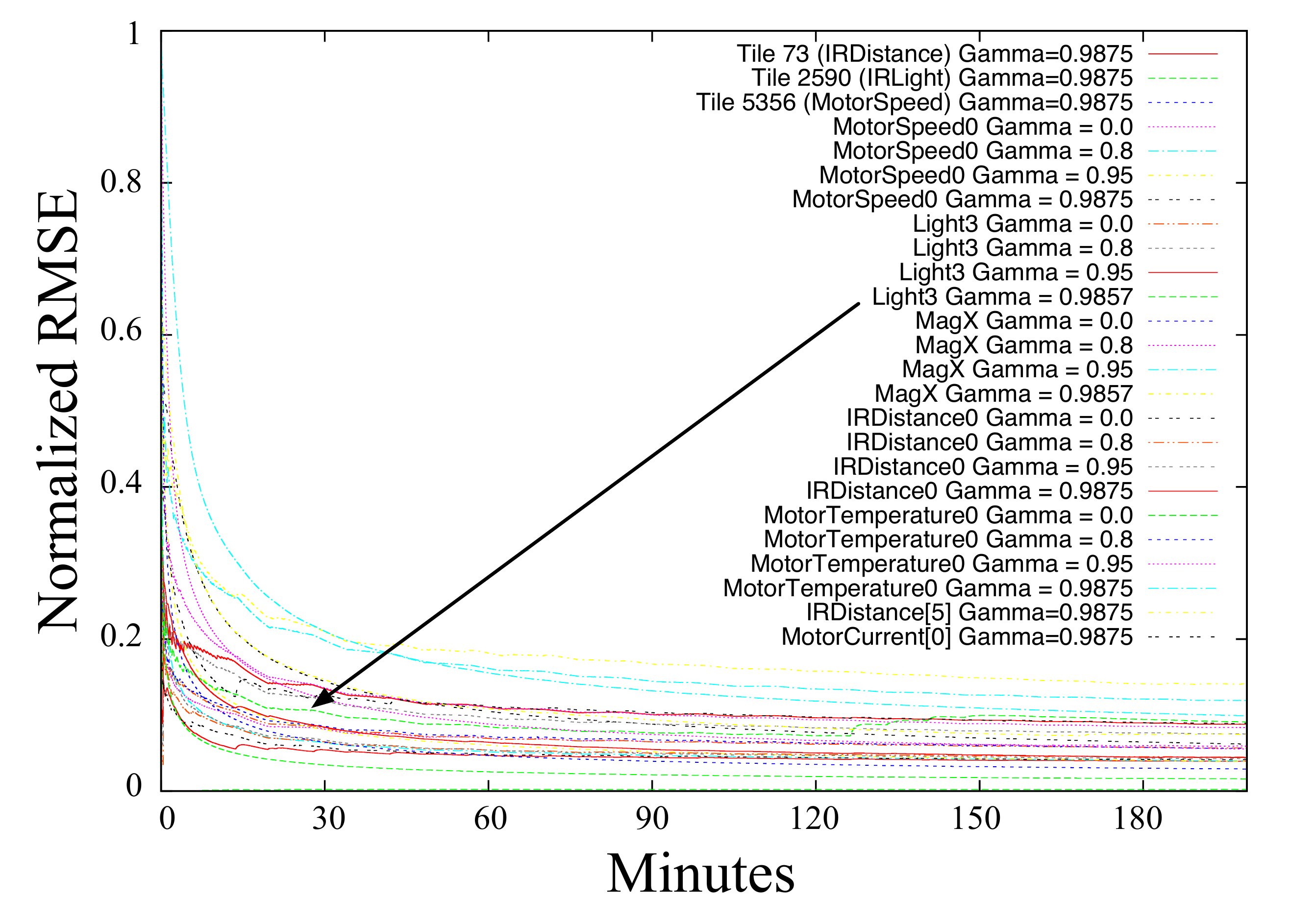}
\caption{\label{learningCurve} Nexting learning curves for the 8-second light sensor predictions (left) and for a representative sample of the TD($\lambda$) predictions (right). Predictions at different time scales have had their root mean squared error (RMSE) normalized by $\frac{1}{1-\gamma^i}$.  The graph on the left is a comparison of different learning algorithms. The jog in the middle of the first graph occurs when the robot stops by the light to cool off its motors, causing the online learners to start making poor predictions.  In spite of the unusual event, the TD($\lambda$) solution still approaches the offline-optimal solution.  TD($\lambda$) performs similarly to a supervised learner TD(1), and they both slightly outperform TD(0).  The curve for the bias unit shows the poor performance of a learner with a trivial representation. The graph on the right shows that seemingly all the TD($\lambda$) predictions are learning well with a single feature representation and a single set of learning parameters.
}
\end{figure}

Having demonstrated that accurate prediction is possible, we now
consider the rate of learning in Figure~\ref{learningCurve}.  The
graphs shows that learning is fast in terms of data (despite the large number of features), 
converging to solutions with low error in the familiar exponential way.
This result is important as is demonstrates that learning
online in real time is possible on robots with a few hours of
experience, even with a large distributed representation.  For
contrast, we also show the learning curve for a trivial representation
consisting only of a bias unit (the single feature that is always 1).  
The comparison serves to highlight that
large informative feature sets are beneficial.  The comparison to the predictive
performance of the offline-optimal solution shows  a vanishing performance gap
by the end of the experiment. 
The second panel of the figure shows a similar pattern of decreasing errors for a sample of the 2160 TD($\lambda$) predictions, showing
that learning many predictions in parallel yields similar results.
A noteworthy result
is that the same learning parameters and representation suffice for
learning answers to a wide variety of nexting predictions without any
convergence problems.  Although the answers continue to improve over
time, the most dramatic gains were achieved after 30 minutes of real time.

\comment{
\begin{figure}
\includegraphics[width=.5\columnwidth]{manyLearningCurves}
\caption{\label{learningCurves} The learning curves for a sample of
  the 2160 nexting predictions shows similar learning performance.  The use of normalized RMSE enables predictions with
  different values of $\gamma$ to be easily compared.  No learner is
  diverging and all the answers are being learned with the same
  parameters and feature representation. }
\end{figure}
}

\section{Discussion}

These results provide evidence that online learning of thousands of
nexting predictions on a robot in parallel is possible, practical, and
accurate.  Moreover, the predictive accuracy is reasonable with just a few
hours of robot experience, no tuning of algorithm parameters, and using a single feature representation for all predictions.  The parallel scalability 
of knowledge-acquisition in this approach is substantially novel
when compared with the predominately sequential existing approaches
common for robot learning.  These results also show that online
methods can be competitive in accuracy with an offline optimization of
mean squared error.

The ease with which a simple reinforcement learning algorithm enables
nexting on a robot is somewhat surprising.  Although the formal theories of reinforcement learning  sometimes give
mathematical guarantees of convergence, there is little guidance for
the choice of features for a task, for selecting learning parameters
across a range of tasks, or for how much experience is required before
a reinforcement learning system will approach convergence.  The experiments show that
we can use the same features across a range of tasks, anticipate events
before they occur, and achieve predictive accuracy approaching that of an
offline-optimal solution with a limited amount of robot experience.

\comment{Both a premise and strength of the nexting approach is its strong grounding in empirical data.

The knowledge acquired by nexting has the benefit of being
adapted to the robot's experience and being fundamentally empirical.  The
target of the learning algorithms is determined solely by the agent's
feature representation, and it is not constrained by assumptions of an
underlying generative process.  In spite of the lack of a model, this
approach still enables offline comparisons with an empirical return
and an optimal offline solution.  As such it provides practitioners
with insight into whether additional experience or additional features
would be the most beneficial for improving predictive performance.
}

\comment{It is also worth reflecting on how this approach to nexting on a robot
compares to the examples of nexting in animals as mentioned in the
introduction.  The fact is that the robot has learned sensory
contingencies across multiple timescales, and has learned it under
learning constraints that are realistic for an animal (online,
real-time, multi-timescale).  %
}

\section{More General Nexting}

The  exponentially discounted predictions that we have focused on in this paper constitute the simplest kind of nexting.  
They are a natural first kind of predictive knowledge to be learned.
  Online TD-style
algorithms can be extended to handle a much broader set of predictions, including time-varying
choices of $\gamma$, time-varying $\lambda$, and even off-policy
prediction (Maei \& Sutton 2010).  It has even been proposed that all world knowledge can be represented by a sufficiently large and diverse set of predictions (Sutton 2009).

\begin{wrapfigure}{r}{.5\columnwidth}
\vspace*{-15pt}
\includegraphics[width=.53\columnwidth]{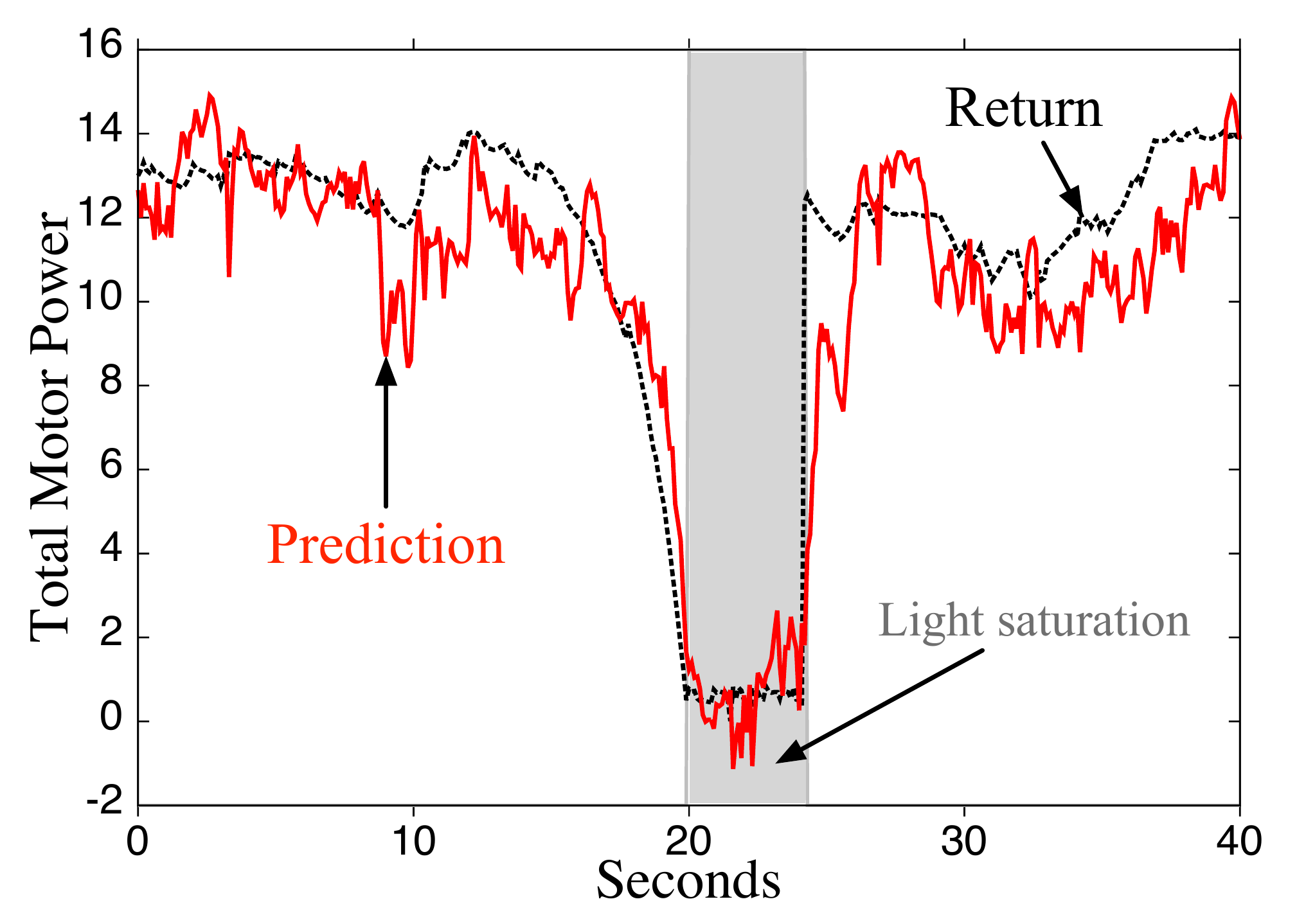}
\vspace*{-15pt}
\caption{\label{fig:Power} Nexting can be extended, for example to
  consider time-varying gamma to predict of the amount of power that
  the robot will expend before a probabilistic pseudo-termination with
  a 2-second time horizon or a saturation event on the light sensor.}
\vspace*{-10pt}
\end{wrapfigure}

As one example of such an extension, consider allowing the discount
rate $\gamma^i$ to vary as a function of the agent's state.  The
algorithmic modifications required are straightforward.  In the
definition of the return in Equation 1, $(\gamma^i)^k$ is replaced
with $\Pi_{j=0}^k \gamma^i_{t+j}$.  In Equation 3, $\gamma^i$ is
replaced with $\gamma^i_{t+1}$ and finally, in Equation 4, $\gamma^i$
is replaced with $\gamma^i_{t}$.  Using the modified definitions, the
robot can predict how much motor power it will consume until either
the light sensor is saturated or approximately two seconds elapse.  This
prediction can be formalized by setting the prediction's target
signal %
to be the sum of instantaneous power consumption of each wheel,
($r=\sum_{i=1}^3 \mbox{MotorVoltage}i \times \mbox{MotorCurrent}i$)
and throttling gamma when the light sensor is saturated ($\gamma^i_t=0.1$ when
the light sensor is saturated and $0.95$ otherwise).  The plots in
Figure~\ref{fig:Power} shows that the robot has learned to anticipate
how much power will be expended prior to reach the light or
spontaneously terminating.

\comment{
Our work provides a novel methodology for evaluating prediction accuracy on a robot. We compare learned predictions with empirical returns and offline-optimal predictions, averaging over salient light observations. Our approach, therefore, provides a practical measure of accuracy without state re-visitation or time consuming evaluation runs--two commonly exploited but limiting assumptions for studying reinforcement learning systems.
}

\section{Conclusions}

We have demonstrated multi-timescale nexting on a physical robot; thousands of
anticipatory predictions at various time-scales can be learned in
parallel on a physical robot in real-time using a reinforcement
learning methodology.  This approach uses a large feature
representation with an online learning algorithm to provide an
efficient means for making parallel predictions.  The algorithms are
capable of making real-time predictions about the future of the
robot's sensors at multiple time-scales using the computational
horsepower of a laptop.  Finally, and key to the practical application of our approach, we have shown that a single feature representation and a single set of learning parameters are sufficient for learning many diverse predictions.  A natural direction for future work would be to extend these results to more general predictions and to \emph{}control.

\section*{Acknowledgements}

The authors thank Mike Sokolski for creating the Critterbot and Patrick Pilarski and Thomas Degris for preparation of Figure 1 and for essential assistance with the experiment briefly reported in Section 6.
This work was supported by grants from Alberta Innovates -- Technology Futures, the National Science and Engineering Reseach Council of Canada, and the Alberta Innovates Centre for Machine Learning.

\section*{References}
\parindent=0pt
\def\hangin{\hangindent=0.15in}
\parskip=6pt

\hangin
Brogden, W. (1939).
\newblock Sensory pre-conditioning.
\newblock {\em Journal of Experimental Psychology 25}(4):323--332.

\hangin
Butz, M., Sigaud, O., G{\'e}rard, P., Eds. (2003).
{\em Anticipatory Behaviour in Adaptive Learning Systems: Foundations, Theories, and Systems}, LNAI 2684, Springer.

\hangin
Carlsson, K., Petrovic, P., Skare, S., Petersson, K.,  Ingvar, M. (2000).
\newblock Tickling expectations: neural processing in anticipation of a sensory
  stimulus.
\newblock {\em Journal of Cognitive Neuroscience 12}(4):691--703.

\hangin
Clark, A. (in press).
Whatever Next? Predictive Brains, Situated Agents, and the Future of Cognitive Science. {\em Behavioral and Brain Sciences}.

\hangin
Dayan, P.,  Hinton, G. (1993).
\newblock Feudal reinforcement learning.
\newblock {\em Advances in Neural Information Processing Systems 5}, pp.~271--278.

\hangin
Gilbert, D. (2006).
\newblock {\em Stumbling on Happiness}.
\newblock Knopf Press.

\hangin
Grush, R. (2004).
\newblock The emulation theory of representation: motor control, imagery, and
  perception.
\newblock {\em Behavioural and Brain Sciences 27}:377--442.

\hangin
Hawkins, J.,  Blakeslee, S. (2004).
\newblock {\em On Intelligence}.
\newblock Times Books.

\hangin
Huron, D. (2006).
\newblock {\em Sweet anticipation: Music and the Psychology of Expectation}.
\newblock MIT Press.

\hangin
Kaelbling, L. (1993).
\newblock Learning to achieve goals.
\newblock {\em In Proceedings of International Joint Conference on Artificial Intelligence}.

\hangin
Levitin, D. (2006).
\newblock {\em This is Your Brain on Music}.
\newblock Dutton Books.

\hangin
Pavlov, I. (1927).
\newblock {\em Conditioned Reflexes: An Investigations of the Physiological
  Activity of the Cerebral Cortex}, translated and edited by G.~V.~Anrep.
\newblock Oxford University Press.

\hangin
Pezzulo, G. (2008).
\newblock Coordinating with the future: The anticipatory nature of
  representation.
\newblock {\em Minds and Machines 18}(2):179--225.

\hangin
Rescorla, R. (1980).
\newblock Simultaneous and successive associations in sensory preconditioning.
\newblock {\em Journal of Experimental Psychology: Animal Behavior Processes
  6}(3):207--216.

\hangin
Singh, S. (1992).
\newblock Reinforcement learning with a hierarchy of abstract models.
\newblock {\em Proceedings of the Conference of the Association for the Advancement of Artificial Intelligence} (AAAI-92), pp.~202--207.

\hangin
Sutton, R.~S. (1988).
\newblock Learning to predict by the method of temporal differences.
\newblock {\em Machine Learning 3}:9--44.

\hangin
Sutton, R.~S. (1990).
\newblock Integrated architectures for learning, planning, and reacting based on approximating dynamic programming,
\newblock  {\em Proceedings of the Seventh International Conference on Machine Learning}, pp.~216--224.

\hangin
Sutton, R.~S. (1995).
\newblock {TD} models: Modeling the world at a mixture of time scales.
\newblock {\em Proceedings of the International Conference on Machine Learning}, pp.~531--539.

\hangin
Sutton, R.~S. (2009). 
The grand challenge of predictive empirical abstract knowledge. 
In: {\em Working Notes of the IJCAI-09 Workshop on Grand Challenges for Reasoning from Experiences.}

\hangin
Sutton, R.~S.,  Barto, A.~G. (1990).
\newblock  Time-derivative models of {P}avlovian reinforcement. 
\newblock In {\em Learning and Computational Neuroscience: Foundations of Adaptive Networks}, pp.~497--537.
\newblock MIT Press.

\hangin
Sutton, R.~S.,  Barto, A.~G. (1998).
\newblock {\em Reinforcement Learning: An Introduction}.
\newblock {MIT} Press.

\hangin
Sutton, R.~S., Modayil, J., Delp, M., Degris, T., Pilarski, P.~M., White, A.,
   Precup, D. (2011).
\newblock Horde: A scalable real-time architecture for learning knowledge from
  unsupervised sensorimotor interaction.
\newblock {\em Proceedings of the 10th International Conference on Autonomous Agents and Multiagent Systems}, pp.~761--768.

\hangin
Sutton, R.~S., Precup, D.,  Singh, S. (1999).
\newblock Between {MDP}s and semi-{MDP}s: A framework for temporal abstraction
  in reinforcement learning.
\newblock {\em Artificial Intelligence 112}:181--211.

\hangin
Tolman, E.~C. (1951).
\newblock {\em Purposive Behavior in Animals and Men}.
\newblock University of California Press.

\hangin
Wolpert, D., Ghahramani, Z.,  Jordan, M. (1995).
\newblock An internal model for sensori-motor integration.
\newblock {\em Science 269}(5232):1880--1882.

\end{document}

%% file: symbols.tex
\newdimen\pHeight
\newdimen\pLower
\newdimen\pLineWidth
\newdimen\pKern
\newdimen\pIR
\newsavebox{\Cbox}
\newsavebox{\vertCmplx}
\newdimen\Cheight
\newdimen\Cwidth
\sbox{\Cbox}{\rm C}
\Cheight=\ht\Cbox
\Cwidth=\wd\Cbox
\advance\Cheight by \pHeight
\sbox{\vertCmplx}{\rule[\pLower]{\pLineWidth}{\Cheight}}
\sbox{\Cbox}{\usebox{\Cbox}\kern\pKern\usebox{\vertCmplx}}
\wd\Cbox=\Cwidth
\def\Re{\mathbb{R}}

\def\Nat{{\rm I\kern\pIR N}}

\def\vec0{{\boldsymbol{0}}}

\newcommand{\beq}{\begin{equation}}
\newcommand{\eeq}{\end{equation}}
\newcommand{\beqa}{\begin{eqnarray}}
\newcommand{\eeqa}{\end{eqnarray}}
\newcommand{\beqan}{\begin{eqnarray*}}
\newcommand{\eeqan}{\end{eqnarray*}}
\newcommand{\ben}{\begin{eqnarray*}}
\newcommand{\een}{\end{eqnarray*}}